\documentclass[letterpaper]{article} 
\usepackage{aaai25}  
\usepackage{times}  
\usepackage{helvet}  
\usepackage{courier}  
\usepackage[hyphens]{url}  
\usepackage{graphicx} 
\usepackage{natbib}  
\usepackage{caption} 
\usepackage{algorithm}
\usepackage{algorithmic}
\usepackage{colortbl}
\usepackage{times}
\usepackage{soul}
\usepackage{url}
\usepackage[utf8]{inputenc}
\usepackage{graphicx}
\usepackage{arydshln}
\usepackage{amsmath}
\usepackage{amsthm}
\usepackage{booktabs}
\usepackage{algorithm}
\usepackage{algorithmic}

\usepackage{subfig}
\usepackage{tcolorbox}
\usepackage{multirow}
\usepackage{enumitem}
\usepackage{makecell}
\usepackage{threeparttable}
\usepackage{amsfonts,mathtools}
\usepackage{mathrsfs}
\usepackage{float}

\newcommand{\bluefont}[1]{ {\color{blue}{#1}}}
\newcommand{\redfont}[1]{{\textcolor{red}{#1}}}

\definecolor{mygreen}{HTML}{00B050}
\newcommand{\greenfont}[1]{{\textcolor{mygreen}{#1}}}

\definecolor{myorange}{HTML}{ED7D31}
\newcommand{\orangefont}[1]{{\textcolor{myorange}{#1}}}

\newcommand{\answerTODO}[1][]{\textcolor{red}{\bf [TODO]}}
\newcommand{\justificationTODO}[1][]{\textcolor{red}{\bf [TODO]}}

\usepackage{newfloat}
\usepackage{listings}
\DeclareCaptionStyle{ruled}{labelfont=normalfont,labelsep=colon,strut=off} 
\floatstyle{ruled}
\newfloat{listing}{tb}{lst}{}
\floatname{listing}{Listing}
\nocopyright

\title{KnowPO: Knowledge-aware Preference Optimization for Controllable \\ Knowledge Selection in Retrieval-Augmented Language Models}
\author{
    Ruizhe Zhang\equalcontrib, Yongxin Xu\equalcontrib,\\ Yuzhen Xiao, Runchuan Zhu, Xinke Jiang, \\
    Xu Chu\textsuperscript{\textdagger}, Junfeng Zhao\footnote{Corresponding authors.}\footnote{Junfeng Zhao is also at the Big Data Technology Research Center, Nanhu Laboratory, 314002, Jiaxing.}, Yasha Wang\textsuperscript{\textdagger}
}
\affiliations{
    Key Laboratory of High Confidence Software Technologies (Peking University) \\ Ministry of Education; School of Computer Science,\\ Peking University, Peking, China\\
}

\begin{document}

\maketitle

\begin{abstract}
By integrating external knowledge, Retrieval-Augmented Generation (RAG) has become an effective strategy for mitigating the hallucination problems that large language models (LLMs) encounter when dealing with knowledge-intensive tasks.
However, in the process of integrating external non-parametric supporting evidence with internal parametric knowledge, inevitable knowledge conflicts may arise, leading to confusion in the model's responses.
To enhance the knowledge selection of LLMs in various contexts, some research has focused on refining their behavior patterns through instruction-tuning.
Nonetheless, due to the absence of explicit negative signals and comparative objectives, models fine-tuned in this manner may still exhibit undesirable behaviors such as contextual ignorance and contextual overinclusion.
To this end, we propose a \textbf{\underline{Know}}ledge-aware \textbf{\underline{P}}reference \textbf{\underline{O}}ptimization strategy, dubbed KnowPO, aimed at achieving adaptive knowledge selection based on contextual relevance in real retrieval scenarios.
Concretely, we proposed a general paradigm for constructing knowledge conflict datasets, which comprehensively cover various error types and learn how to avoid these negative signals through preference optimization methods. Simultaneously, we proposed a rewriting strategy and data ratio optimization strategy to address preference imbalances.
Experimental results show that KnowPO outperforms previous methods for handling knowledge conflicts by over 37\%, while also exhibiting robust generalization across various out-of-distribution datasets.
\end{abstract}

\section{I. Introduction}
\label{introduction}
Large Language Models (LLMs)~\cite{taylor2022galactica, zhao2023survey} have been widely applied in various fields, such as natural language processing, question-answering systems, and text generation, giving rise to numerous AI applications~\cite{kaplan2020scaling, vu2024gptvoicetasker}. These models exhibit outstanding performance in many tasks, primarily due to their large-scale parameters and extensive pre-training data~\cite{ziegler2020finetuning, wang2023selfinstruct}. However, because of the static nature of the training data, LLMs may generate seemingly coherent but actually unreliable information, a phenomenon known as ``hallucination"~\cite{ji2023survey,cao-etal-2020-factual,10.1145/3571730}, due to outdated knowledge and long-tail knowledge~\cite{he2022rethinking,kandpal2023large,jiang2024hykgehypothesisknowledgegraph}. Retrieval-Augmented Generation (RAG) paradigm~\cite{izacard2022atlas,asai2023self,asai2023retrieval}, within a retrieve-and-read framework, leverages information from reliable knowledge bases to compensate the static nature of the Internal knowledge of LLM. However, the performance of RAG framework is limited by the \textbf{knowledge conflicts} between internal knowledge stored in LLM parameters and external database~\cite{xu2024knowledgeconflictsllmssurvey,jin2024tugofwarknowledgeexploringresolving}. In this paper, we focus on resolving knowledge conflict by adhere to the retrieved knowledge, and meanwhile, improving the robustness against noise in the retrieved context.

\begin{figure*}[t]
  \centering
  \vspace{-0.8cm}
\includegraphics[height=0.4\textwidth]{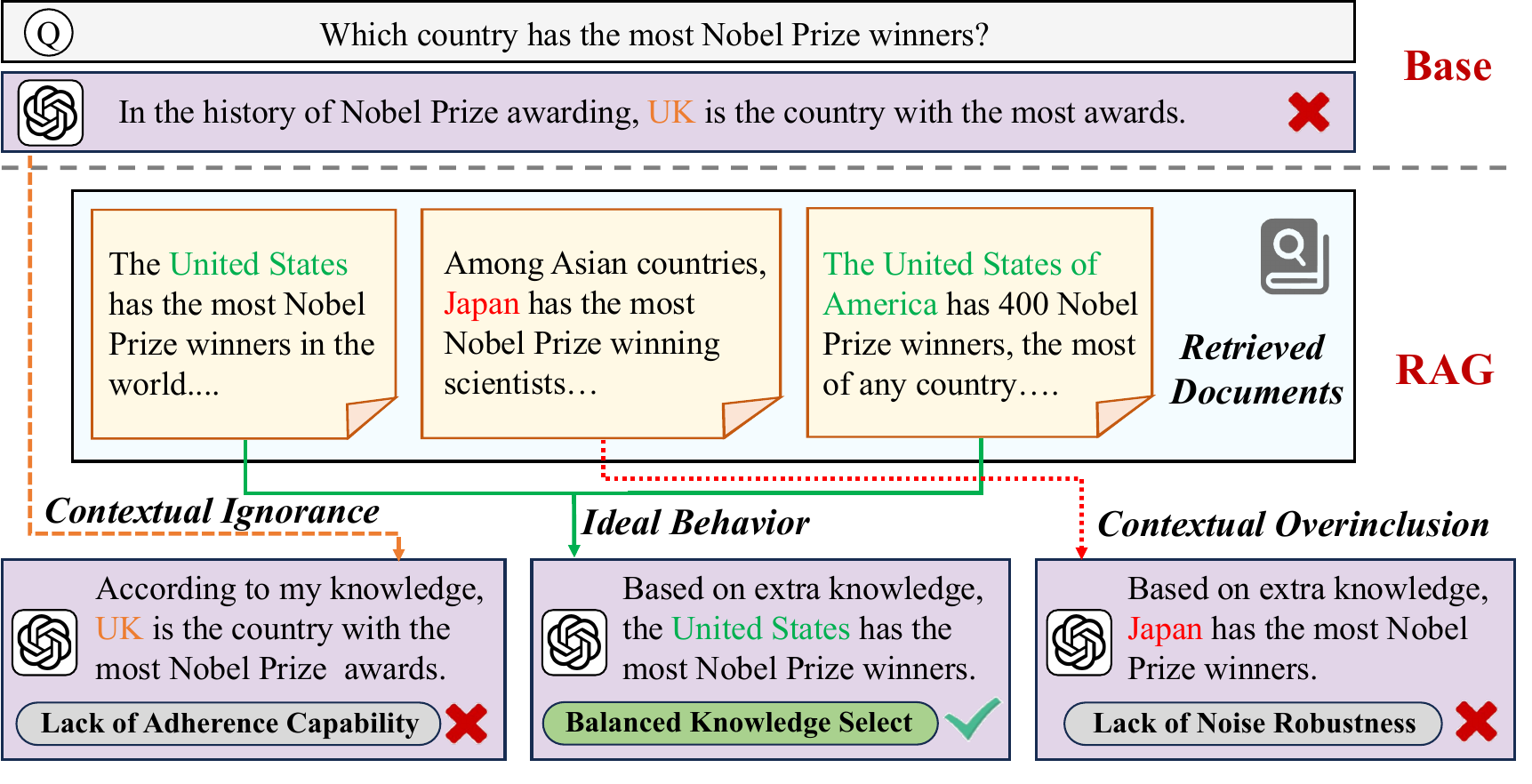}
  \vspace{-0.2cm}
  \caption{An illustrative example of how does LLM behave when encountering knowledge conflicts in RAG scenarios. }
  \vspace{-0.25cm}
  \label{fig:intro}
\end{figure*}

In response to the aforementioned issue, a mainstream approach is to construct specific instruction-tuning datasets to optimize the knowledge prioritization of LLMs in contexts with varying degrees of relevance~\cite{li2022largelanguagemodelscontrollable,xue2023improvingfactualconsistencyknowledgegrounded}. However, as shown in Figure \ref{fig:intro}, achieving a balance between \textbf{adherence capability} and \textbf{noise robustness} is highly challenging. On one hand, when the LLM heavily relies on external knowledge, it risks over-focusing on irrelevant retrieval contexts, struggling to effectively discern noise. On the other hand, an excessive emphasis on enhancing the LLM’s noise resistance can inadvertently filter out useful contextual information~\cite{wu2024clashevalquantifyingtugofwarllms}. Moreover, the manifestation of these capabilities is closely related to the complexity of the context in real-world RAG scenarios~\cite{longpre2022entitybasedknowledgeconflictsquestion,xie2024adaptivechameleonstubbornsloth}. Therefore, it is crucial to address the balance between adherence capability and noise robustness in real RAG scenarios.

Therefore, our insights stem from the error types observed in real-world scenarios involving RAG. We observe that existing research fails to distinguish between supervisory signals, leading to adherence capability and noise robustness being treated as analogous instruction-following positive examples. This leads to contradictory signals during instruction-following training, causing learning variance and impeding the effective acquisition of both capabilities. To address this, we propose a more nuanced optimization approach that introduces preference data specifically describing adherence capability and noise robustness. Leveraging efficacious and extensively utilized Direct Preference Optimization (DPO)~\cite{rafailov2024directpreferenceoptimizationlanguage}, we optimize the model's ability to leverage external knowledge, thereby enhancing the overall efficacy of RAG.

Although seemingly straightforward, implementing this intuition faces these challenges: \textbf{(C1)} How to more accurately simulate complicated context in real-world RAG scenarios and introduce more comprehensive, fine-grained negative signals? \textbf{(C2)} How to resolve data discrepancies in preference learning to avoid behavior pattern imbalances?

By jointly considering the above issues,  we propose KnowPO, a \textbf{\underline{Know}}ledge-aware \textbf{\underline{P}}reference \textbf{\underline{O}}ptimization strategy, which constructs comprehensive and balanced preference relations to optimize LLMs' knowledge selection in different contexts. 
{\textbf{i)}} Specifically, we simulated real-world RAG scenarios at the input level. We perform refined noise classification based on the relevance between the knowledge context and the question topic, and explore combination methods with evidence to form conflicting context and irrelevant context. At the output level, we simulate two common error types in different context relevance scenarios: \textbf{Contextual Ignorance} and \textbf{Contextual Overinclusion}, and develop training strategies to avoid these errors. 
{\textbf{ii)}} Secondly, we propose a rewriting strategy to address length imbalance and a data ratio balancing strategy to address behavior pattern imbalance, using DPO to optimize LLMs' adherence capability and noise robustness. These strategies not only eliminate length biases and imbalances in behavior pattern distribution but also enhance the exhaustiveness of the model's responses. This prevents degradation of conversational abilities that can occur when training on datasets with shorter answers, such as in reading comprehension tasks.

Our main contributions are summarized as follows:
\begin{itemize} [leftmargin=*]
    \item We observed that LLMs in RAG scenarios fail to effectively balance adherence capability and noise robustness. and thus proposed the KnowPO framework, which refines negative supervisory signals to enhance LLM behavior when encountering knowledge conflicts.
    \item We proposed a general paradigm for constructing knowledge conflict datasets, comprehensively covering various error types and generalizable to different model architectures. We also proposed a rewriting strategy and data ratio optimization strategy to address preference imbalances.
    \item We validated our method's training effectiveness on multiple models and datasets and tested its generalization ability in out-of-distribution (OOD) scenarios. The results indicate that our method not only improves the performance of models on test sets but also enhances their adaptability and robustness when confronted with unknown data.
\end{itemize}

\section{II. Related Work}
\label{related work}
\paragraph{\textbf{Knowledge Conflicts. }} 
Numerous studies have explored LLMs' behavior in knowledge conflict scenarios, providing valuable insights for our work. ~\citet{longpre2022entitybasedknowledgeconflictsquestion} discovered that large Pre-trained Language Models often prefer parametric knowledge over contextual information when facing knowledge conflicts. Recently, with the emergence of LLM such as ChatGPT~\cite{OpenAI2023GPT4TR} and PaLM~\cite{chowdhery2022palm}, researchers re-examined this issue. ~\citet{wu2024clashevalquantifyingtugofwarllms} highlighted that this tendency to disregard context is influenced by the model's prior token probability. Specifically, parametric knowledge with higher prior probabilities is more challenging to update to conflicting knowledge. ~\citet{Kassner_Schütze_2019} demonstrated that LLMs are susceptible to being misled by task-irrelevant context. Furthermore, ~\citet{tan2024blindedgeneratedcontextslanguage} indicated that the model's contextual preferences are linked to the semantic completeness of the context and its relevance to the question.

Several studies aim to improve the adherence of LLMs to context amid knowledge conflicts. For instance, Knowledge Aware Fine-Tuning (KAFT)~\cite{li2022largelanguagemodelscontrollable} enhances models' ability to use external knowledge by creating challenging counterfactual knowledge from training datasets and incorporating irrelevant knowledge to boost noise resistance. However, as previously mentioned, the applicability of this approach in real-world RAG scenarios is limited. Additionally, decoding-based methods~\cite{jin2024tugofwarknowledgeexploringresolving,chen2022richknowledgesourcesbring}, like Context-Aware Decoding (CAD)~\cite{shi2023trustingevidencehallucinatecontextaware}, adjust LLMs' output probabilities during token generation, akin to contrastive decoding, conditioned on relevant context. However, this approach may impact the semantic coherence of long responses. Moreover, prompt-based methods employ sophisticated designed prompts to ensure that LLMs adhere to the provided context~\cite{si2023promptinggpt3reliable,zhou2023contextfaithfulpromptinglargelanguage}. However, research shows that merely modifying prompts doesn't significantly alter LLMs' internal prior token probabilities~\cite{wu2024clashevalquantifyingtugofwarllms}, potentially limiting the effectiveness of this approach.

\paragraph{\textbf{Retrieval-Augmented Generation. }}
RAG incorporates the external knowledge retrieval component via prompt engineering to achieve more factual consistency, enhancing the reliability and interpretability of LLMs' responses~\cite{gao2024RAGsurvey,yu2023generateretrievelargelanguage,lewis2021retrievalaugmented}. Some studies have made improvements to the \textit{retrieve-and-read} framework by generating intermediate contexts using the parameter knowledge acquired during the pretraining phase, thereby enhancing the quality of the final response. These intermediate contexts may include commonsense knowledge~\cite{liu2022generatedknowledgepromptingcommonsense}, domain-specific knowledge~\cite{luo2023augmentedlargelanguagemodels}, and chain-of-thought(COT) reasoning process~\cite{wei2023chainofthought,kojima2023largelanguagemodelszeroshot,li2023chainofknowledge}. Furthermore, ~\citet{zhao2023verifyandeditknowledgeenhancedchainofthoughtframework}, ~\citet{wang2023knowledgedrivencotexploringfaithful} and ~\citet{yu2023chainofnote} have utilized retrieved knowledge to edit the intermediate contexts of the COT process, thereby updating conflicting knowledge. However, these intermediate contexts generated by LLMs may contain hallucinations, potentially misleading the retrieval or reader models. Additionally, the frequent interactions with LLMs result in inefficiencies in real-world applications~\cite{jiang2024hykgehypothesisknowledgegraph}.

\paragraph{\textbf{Knowledge Editing. }} Knowledge editing is a classic method for updating model knowledge, focusing on identifying how models store factual knowledge and designing effective strategies to update parametric knowledge stored in pre-trained models~\cite{decao2021editingfactualknowledgelanguage,onoe2023lmslearnnewentities,meng2023locatingeditingfactualassociations}. ~\citet{jang2022continualknowledgelearninglanguage} proposed a continual learning framework aimed at updating outdated knowledge while preserving stable knowledge that is unaffected by temporal changes. However, these strategies may unintentionally affect unrelated parameters or cause inconsistencies within the model's internal knowledge~\cite{pinter2023emptyingoceanspoonedit,xu2024knowledgeconflictsllmssurvey,Wang_2023}. Moreover, in the constantly evolving context of RAG scenarios, the sequential training method for injecting new knowledge proves impractical.

\section{III. Methodology}
\label{Methodology}

\subsection{A. Task Definition}
\label{sec: Task Definition}
Given an LLM $\Theta$ and an input natural language question $q$, we ask $\Theta$ to generate a response $\alpha = \Theta(q)$, representing the parametric knowledge for $q$. Assume in a typical \textit{retrieve-and-read} framework, context $\tau$ is a permutation of $D_j^r, j = 1,2,\ldots,K$, which represents a set of documents retrieved based on $q$. And $S = \{a_i\}, i = 1,2,\ldots,N$ constitutes the set of contextual answer, each of which is derived from a retrieved document $D_{\tau_i}^r$. We can simplify the RAG task into $y = \Theta(q \| \tau)$, where $y$ is output of $\Theta$ based on context $\tau$. Note that $K$ is not necessarily equals with $N$, because some retrieved documents may not contain any answer for $q$ and are known as noises. 

 It's clearly that $\alpha$ and $a_i$ are independent. \textbf{Knowledge conflict} appears when $\alpha \notin S$, and at this time response $y$ of $\Theta(q \| \tau)$ can be uncertain. To simplify the discussion, we limit $N$ to a maximum of $1$, which means context $\tau$ contains at most one document $D_{\epsilon}^r$ from which the answer can be derived. Our purpose is to make sure $y=a_\epsilon$ when $|S|=1$ and $y=\alpha$ when $|S|=0$. In other word, LLM $\Theta$ should use appropriate external knowledge when there exists a document which contains the necessary knowledge regardless of conflicting with parameter knowledge, while use its parameter knowledge when retrieved documents are all irrelevant.

\subsection{B. Contradictory Knowledge}
\label{sec: Contradictory Knowledge}
Constructing knowledge that conflicts with LLM's parameter knowledge is crucial to condition $|S|=1$. For question $q$ in RAG scenarios, this conflict is reflected in conflicting answers $a_{cf}$ which are inconsistent with LLM's parameter answer $\alpha$. It is important to note that these conflicting answers $a_{cf}$ do not necessarily have to be correct, nor is the LLM's parameter answer $\alpha$ always incorrect. In our approach, both answers can be incorrect to the question as long as they conflict with each other. The key to knowledge conflict lies in the conflict itself, regardless of correctness. This addresses a common misconception in previous work, where researchers often ensured that one answer was correct and the other incorrect~\cite{tan2024blindedgeneratedcontextslanguage,wu2024clashevalquantifyingtugofwarllms}, which not only increased the difficulty of data filtering but also overlooked some knowledge conflict scenarios.

Specifically, we first extract world knowledge acquired during the pretraining phase of the large model, marked as parameter answer $\alpha$. We encourage LLM to abstain from answering when uncertain. Additionally, we refine the response formats for other parametric knowledge. The revised results are presented in Table \ref{Dataset Examples}.

For a given question $q$ and LLM's parameter answer $\alpha$, there are two potential sources of conflicting answers $a_{cf}$. The first is the realistic answer $a_{real}$ to the question. The second is a fabricated answer $a_{ctf}$ generated using GPT-4 that deviates from the realistic answer $a_{real}$. The latter is often referred to as a counterfactual answer, which we require to be as plausible as possible. Thus, for a question $q$ and LLM's parameter answer $\alpha$, we can obtain at least one conflicting answer, ensuring it is not overly far-fetched. The few-shot prompts can be found in Appendix A.

\subsection{C. Context Formulation} 
\label{sec: Context Formulation}
In this section, we illustrated how to formulate context $\tau$ based on different kinds of knowledge conflict.

To align with the RAG scenario, we utilized the SQuAD2.0 dataset~\cite{rajpurkar2018knowdontknowunanswerable}, a reading comprehension dataset encompassing multiple general domains, with a substantial corpus of documents and associated QA tasks. Notably, besides corpora collected from Wikipedia, SQuAD2.0 is also annotated by humans to determine whether a document can yield an answer for a specific question. Previous research has highlighted that treating a relevant yet non-informative document as a reference external knowledge source can impair LLM's adherence capabilities~\cite{li2022largelanguagemodelscontrollable}. Following the chunk-size commonly used in RAG tasks~\cite{shi2023largelanguagemodelseasily}, we set the length of context $\tau$ to $K=4$.

For scenarios with $|S|=1$, we initially select pertinent documents from SQuAD2.0 based on the conflicting knowledge:  For question $q$ and realistic answer $a_{real}$, we directly select the corresponding document $D_{\epsilon}^r$ from the original dataset; and for question $q$ and counterfactual answer $a_{ctf}$, we replace all occurrences of $a_{real}$ with $a_{ctf}$ in $D_{\epsilon}^r$. Subsequently, we select one relevant document on the same topic and two documents on different topics based on semantic similarity. We ensure that these three documents are incapable of answering the question $q$. These four documents are then shuffled to constitute the conflicting context $\tau_{cf}$.

For scenarios with $|S| = 0$, we distinguish between hard and easy irrelevant documents. Hard documents, derived from human annotations, consist of two documents that are on related topics but cannot answer the question. Easy documents are randomly selected, consisting of two documents on unrelated topics. These four documents are then shuffled to constitute the irrelevant context $\tau_{ir}$.

\begin{table*}[ht] 
\vspace{-0.4cm}
\centering 
\begin{tabular}{
  |>{\raggedright\arraybackslash}m{2.11cm}
  |>{\centering\arraybackslash}m{1.1cm}
  |>{\raggedright\arraybackslash}m{6.25cm}
  |>{\raggedright\arraybackslash}m{6.8cm}|
}
\hline
\multicolumn{4}{|>{\raggedright\arraybackslash}m{16.26cm}|}{\textit{Sample Question}: Who is the Democratic presidential candidate in the 2024 US presidential election? 
\quad\quad\quad\quad\quad\quad\quad\quad
\textit{Parameter Answer}: \orangefont{The Democratic candidate is Joe Biden.}} \\
\hline
\multicolumn{2}{|c|}{\textbf{}} & \textbf{Conflicting Context} & \textbf{Irrelevant Context} \\
\hline
\multicolumn{2}{|c|}{\textbf{Context Example} }& {With President Joe Biden dropping out of the race on July 21, \greenfont{Vice President Kamala Harris became the presumed Democratic nominee.} The search for her vice presidential running mate is closely watched, with top contenders including Secretary of Transportation \redfont{Pete Buttigieg}, Arizona Senator \redfont{Mark Kelly}, Illinois Governor \redfont{J.B. Pritzker}...} & {...a decision not without precedent in political history. Back in 1968, \redfont{President Lyndon B. Johnson} of the Democratic Party also opted out amidst intense political challenges. Following Johnson's withdrawal, \redfont{Hubert Humphrey} took over as the Democratic nominee for president.} \\
\hline
\multirow{6}{*}{\textbf{Gold Output}} &  Ideal Answer  & \greenfont{The Democratic candidate is Kamala Harris.} & \orangefont{The Democratic candidate is Joe Biden.} \\
\cline{2-4}
& Revised Result & {Based on supplemental knowledge and my own understanding, the answer to this question is that \greenfont{the Democratic candidate is Kamala Harris.}} & {Supplemental knowledge does not answer this question, but based on my knowledge, the answer to this question is that \orangefont{the Democratic candidate is Joe Biden.}} \\

\hline
\multirow{5}{=}{\textbf{Contextual Overinclusion}} & Error Answer & \redfont{The Democratic candidate is Mark Kelly.} & \redfont{The Democratic candidate is Hubert Humphrey.} \\
\cline{2-4}
& Revised Result & {Based on supplemental knowledge and my own understanding, the answer to this question is that \redfont{The Democratic candidate is Mark Kelly.}} & {Based on supplemental knowledge and my own understanding, the answer to this question is that \redfont{The Democratic candidate is Hubert Humphrey.}} \\
\hline
\multirow{6}{=}{\textbf{Contextual Ignorance}} &  Error Answer &  \orangefont{The Democratic candidate is Joe Biden.} & / \\
\cline{2-4}
& Revised Result & {Supplemental knowledge does not answer this question, but based on my knowledge, the answer to this question is that \orangefont{the Democratic candidate is Joe Biden.}} & / \\
\hline
\end{tabular}
\vspace{-0.15cm}
\caption{An example of how the KnowPO dataset is formulated. LLM's parameter knowledge are marked in \orangefont{orange}, while conflicting knowledge in context are marked in \greenfont{green}, and noisy information is presented in \redfont{red}.}
\label{Dataset Examples}
\end{table*}

\subsection{D. Error Type Analyse}
\label{sec: Error Type Analyse}
As previously mentioned, we expect LLMs to utilize contextual knowledge when encountering conflicting context, while relying on parameter knowledge when faced with irrelevant context. These two modes of handling context reflect the model's adherence capability and noise robustness, respectively. In practical RAG scenarios, deficiencies in these capabilities manifest as two distinct error types: one in which the LLM incorrectly uses irrelevant contextual information to construct answers, termed \textbf{Contextual Overinclusion}; and another where the LLM disregards the context entirely and relies exclusively on its parameter knowledge, termed \textbf{Contextual Ignorance}. These errors can occur with both types of contexts as illustrated in Table \ref{Dataset Examples}. To address these issues, we have meticulously designed a dataset comprising positive and negative sample pairs to specifically target and mitigate these errors.

\paragraph{\textbf{Contextual Overinclusion Error. }}
In situations with conflicting contexts, the ideal behavior of the LLM demonstrating adherence capability, as shown by positive samples in Table \ref{Dataset Examples}, is to answer using the conflicting knowledge present in the context. However, when contextual overinclusion occurs, LLM often utilizes inappropriate information from the context due to insufficient noise robustness and contextual understanding capability. For instance, in the example presented in Table \ref{Dataset Examples}, LLM chooses noisy information marked in red. To address this error, we constructed negative samples by using a prompt mechanism to guide GPT-4 to generate incorrect answers from conflicting contexts. To ensure the quality of the generated data, we adhered to stringent validation criteria: (1) The generated answers must be derived from the context, ensuring that the error is unequivocally attributable to contextual overinclusion; (2) The generated answers should be as plausible as possible and distinctly different from the conflicting answers, thereby ensuring the high quality of the data. 

In situations with irrelevant contexts, it is evident that positive sample for noise robustness is to use LLM's parametric knowledge to respond. When this error occurs, LLM may fail to recognize the context as irrelevant, leading it to use contextual information instead of disregarding it. Similar to contextual overinclusion in conflicting contexts, we constructed corresponding negative samples by using GPT-4 to extract incorrect answers from irrelevant contexts. The prompts used to generate contextual overinclusion is structured as follows:

\begin{tcolorbox}
[colback=lightgray!20,colframe=darkgray!80,title=Prompt: Generate Contextual Overinclusion]
Please select a word from the provided context as an alternative answer to this question.

Question: \{\textit{Question $q$}\}

Potential answer: \{\textit{Conflicting Answer $a_{cf}$}\}

Context: \{\textit{Context $\tau$}\}\\

Please follow these requirements:

1. The answer must not be the same as the potential answer.

2. The alternative answer does not need to be correct, but it must appear in the context.

3. The alternative answer must be in a form that can answer the question and should be as reasonable as possible.
\end{tcolorbox}

\paragraph{\textbf{Contextual Ignorance Error. }}
Contextual Ignorance occurs when the LLM disregards the context in its response, a behavior deemed erroneous solely in conflicting contexts. During such episodes, LLM may either fail to recognize the utility of the context or, even upon recognizing it, may opt to disregard the conflicting answer in favor of relying on its parameter knowledge. For instance, in the example shown in Table \ref{Dataset Examples}, LLM answers the question without utilizing supplemental knowledge. To simulate this error, we constructed negative samples by extracting LLM's response to the query in the absence of any contextual support, ensuring that the answer aligns with an inappropriate erroneous response. Prompt used can be found in Appendix A.

\subsection{E. Training Method}
\label{sec: Training Method}
Our training consists of two phases. First, we perform instruction tuning using the conflicting knowledge and contexts constructed in {Section III.B} and {Section III.C} to enhance the LLM's adherence capability and noise robustness in RAG task scenarios. Next, we utilize the preference dataset in Section III.D for DPO training to further improve the LLM's ability to avoid the two types of errors, while ensuring that its final responses align with user preferences.

\paragraph{\textbf{Instruction Tuning. }} 
Instruction tuning is a multi-task learning framework that enables the use of human-readable instructions to guide the output of LLMs. Given a source text and task-specific instructions, the model is trained to generate a sequence of tokens representing the desired output structure and its corresponding labels. Reviewing our definition of adherence capability and noise robustness, we would like to get a finetuned model $\Theta_{ft}$ from original LLM $\Theta$ that satisfies the following criteria:
\begin{equation*}
\begin{aligned}
&|S| = 1 : \Theta_{ft}(q \| \tau_{cf}) = a_{cf}, && \text{where } \exists D^{r}_\epsilon \in \tau_{cf}, D^{r}_\epsilon \to a_{cf}\\
&|S| = 0  :\Theta_{ft}(q \| \tau_{ir}) = \alpha, && \text{where }\Theta(q) = \alpha
\label{eq:sft task}
\end{aligned}
\end{equation*}
Note that although the presence of the answer in $\tau$ was distinguished during dataset construction, the LLM does not possess this prior knowledge. The model must independently determine the context type and formulate a response during the RAG task. The instruction prompts we used are outlined in Appendix A.

\paragraph{\textbf{Direct Preference Optimization. }} 
As previously discussed, LLMs may exhibit errors contextual overinclusion and contextual ignorance in real-world RAG scenarios. To further enhance adherence capability and noise robustness, we propose a Knowledge-aware Preference Optimization(KnowPO) training strategy. This strategy employs three types of preferences between positive and negative samples in two different contextual settings to conduct DPO training on the LLM. Details of preference pairs construction can be found in {Section III.D}. Using this approach, we train the LLM to avoid these errors and improve its ability to utilize different contexts.

During preparing data for DPO, we also identified two preference imbalances that impact training effectiveness.

\begin{itemize}[leftmargin=*,noitemsep,topsep=2pt]
    \item \textit{\textbf{Length Imbalance. }} Some studies suggest that reward hacking observed in RLHF can also negatively impact DPO training~\cite{gao2022scalinglawsrewardmodel,park2024disentanglinglengthqualitydirect}. We observed that in our previously constructed dataset, for the same preference pair, the positive sample was often the better-formatted and longer response, while the negative sample was a shorter conflicting answer. Due to the tendency of LLMs to be influenced by length bias during DPO~\cite{singhal2024longwaygoinvestigating}, they might prefer generating longer responses, which overall manifests as a greater tendency to refuse answering rather than providing a conflicting answer. To mitigate this issue, we standardized the format for all positive and negative samples in Table \ref{Dataset Examples}, aligning their lengths to ensure that the average length $len_{win}$ approximately equals $len_{loss}$.
    \item \textit{\textbf{Error Type Imbalance. }} Given that the preference pairs related to error contextual ignorance in conflicting context guide the LLM to ``utilize contextual knowledge without rejecting it'', while the preference pairs associated with error contextual overinclusion in irrelevant context exhibit a tendency towards ``rejecting the use of contextual knowledge'', we realized that the ratio of these two contrasting preference pairs could significantly influence training efficacy. During KnowPO training, we ensured that the proportion $\mathcal{R}_{error}$ of these two types of data was maintained at approximately 1:1. Furthermore, we validated the importance of this ratio $\mathcal{R}_{error}$ in subsequent experiments.
\end{itemize}

\section{IV. Experiments}
\label{Experiments}
\begin{table*}[!t]
\vspace{-0.4cm}
\caption{Performance comparison (in percent) on Squad2.0-Eval, RGB and KNOT. \redfont{Red shading} indicates the best-performing model, while \bluefont{blue} signifies the second-best in the ablation study, and \greenfont{green} signifies the second-best in baselines. }
\vspace{-0.2cm}
\fontsize{9pt}{11pt}\selectfont  
\setlength{\tabcolsep}{1mm}     
\label{tab:comparison}
\centering
\resizebox{\linewidth}{!}{
\begin{tabular}{cc|cc|cc|cc|cc|cc|cc}
\hline
\textbf{LLM Turbo} & \multicolumn{1}{c|}{\textbf{LLM}} & \multicolumn{6}{c|}{Baichuan2-7B-Chat} & \multicolumn{6}{c}{Llama2-13B-Chat} \\
\hline
\multirow{2}{*}{\textbf{Method}} & \multicolumn{1}{c|}{\textbf{Dataset}} & \multicolumn{2}{c|}{Squad2.0-Eval} & \multicolumn{2}{c|}{RGB} & \multicolumn{2}{c|}{KNOT} & \multicolumn{2}{c|}{Squad2.0-Eval} & \multicolumn{2}{c|}{RGB} & \multicolumn{2}{c}{KNOT} \\
\cline{2-14}
& \textbf{Metric} & $R_{Ad}$ & $R_{Ro}$ & $R_{Ad}$ & $R_{Ro}$ &$R_{Ad}$ & $R_{Ro}$ &$R_{Ad}$ & $R_{Ro}$ &$R_{Ad}$ & $R_{Ro}$ & $R_{Ad}$ & $R_{Ro}$   \\
\hline
\multirow{6}{*}{\textbf{Baselines}} & Base & {43.51$\pm$0.87} & {9.80$\pm$1.05} & {65.00$\pm$0.70} & {24.00$\pm$1.61} & {26.42$\pm$0.53} & {7.65$\pm$0.90} & {52.71$\pm$0.85} & {11.95$\pm$0.34} & {69.00$\pm$0.68} & {25.00$\pm$0.85} & {49.66$\pm$0.84} & {21.67$\pm$0.85} \\
& Prompt & {53.74$\pm$0.94} & {8.60$\pm$1.21} & \cellcolor{green!15}{79.50$\pm$0.64} & {19.50$\pm$0.90} & {44.65$\pm$0.70} & {14.51$\pm$0.49} & {60.76$\pm$0.98} & {10.59$\pm$0.30} & \cellcolor{green!15}{73.50$\pm$1.27} & {19.50$\pm$1.30} & {41.14$\pm$1.28} & {22.62$\pm$1.30}\\
& COT & {54.65$\pm$0.98} & {10.20$\pm$0.62} & {77.50$\pm$0.67} & {21.00$\pm$0.62} & {44.13$\pm$1.05} & {15.29$\pm$0.53} & {57.13$\pm$1.07} & {12.85$\pm$0.43} & {70.50$\pm$0.71} & {25.00$\pm$0.89} & {41.06$\pm$0.70} & {23.53$\pm$0.64}\\
& COT-VE & {44.83$\pm$0.68} & {8.41$\pm$1.12} & {66.00$\pm$0.31} & {19.50$\pm$0.99} & {27.71$\pm$0.99} & {13.88$\pm$1.01} & {54.52$\pm$0.59} & {10.17$\pm$0.30} & {70.00$\pm$0.79} & {14.50$\pm$0.70} & {52.33$\pm$0.64} & {18.35$\pm$0.97} \\
& KAFT & \cellcolor{green!15}{58.83$\pm$0.72} & \cellcolor{green!15}{21.43$\pm$0.94} & {75.00$\pm$0.36} & \cellcolor{green!15}{27.00$\pm$0.45} & \cellcolor{green!15}{54.45$\pm$1.08} & \cellcolor{green!15}{17.93$\pm$0.41} & \cellcolor{green!15}{65.73$\pm$0.58} & \cellcolor{green!15}{34.34$\pm$1.28} & \cellcolor{green!15}{73.50$\pm$0.47} & \cellcolor{green!15}{29.50$\pm$0.46} & \cellcolor{green!15}{62.21$\pm$0.51} & \cellcolor{green!15}{24.47$\pm$0.5} \\
& CAD & {35.83$\pm$0.78} & {7.50$\pm$0.97} & {55.50$\pm$0.56} & {22.50$\pm$0.83} & {21.72$\pm$1.05} & {6.99$\pm$1.19} & {41.73$\pm$1.31} & {10.94$\pm$0.57} & {64.50$\pm$1.55} & {23.50$\pm$1.57} & {35.71$\pm$0.98} & {19.96$\pm$0.84}\\
\cdashline{1-14}
\textbf{Ours} & \textbf{KnowPO} & \cellcolor{red!20}{80.64$\pm$0.36} & \cellcolor{red!20}{38.77$\pm$0.42} & \cellcolor{red!20}{93.50$\pm$0.30} & \cellcolor{red!20}{37.00$\pm$0.31} & \cellcolor{red!20}{69.95$\pm$0.41} & \cellcolor{red!20}{39.73$\pm$0.34} & \cellcolor{red!20}{76.11$\pm$0.37} & \cellcolor{red!20}{44.64$\pm$0.30} & \cellcolor{red!20}{83.50$\pm$0.32} & \cellcolor{red!20}{37.50$\pm$0.28} & \cellcolor{red!20}{77.03$\pm$0.33} & \cellcolor{red!20}{38.28$\pm$0.48} \\
\hline
\rowcolor[gray]{0.95}
\multicolumn{2}{c|}{*\textbf{Performance Gain $\uparrow$}} & {37.07$\sim$125.06} & {80.91$\sim$416.93} & {17.61$\sim$68.47} & {37.04$\sim$89.74} & {28.47$\sim$222.05} & {121.58$\sim$468.38} & {15.79$\sim$82.39} & {29.58$\sim$338.94} & {13.61$\sim$29.46} & {27.12$\sim$158.62} & {23.82$\sim$115.71} & {49.94$\sim$108.61} \\
\hline
\multirow{2}{*}{\textbf{Ablation}} & KnowPO (w/o DPO) & \cellcolor{cyan!20}{71.09$\pm$0.30} & {36.50$\pm$0.79} & {89.50$\pm$0.99} & {31.00$\pm$1.01} & {64.45$\pm$0.98} & {36.50$\pm$0.43} & \cellcolor{cyan!20}{75.96$\pm$0.59} & {42.87$\pm$1.06} & {80.00$\pm$0.70} & {35.00$\pm$1.06} & \cellcolor{cyan!20}{70.28$\pm$0.48} & {36.21$\pm$0.64} \\
& KnowPO (w/o SFT) & {69.39$\pm$0.67} & {37.50$\pm$0.79} & \cellcolor{cyan!20}{92.50$\pm$1.47} & {35.00$\pm$0.58} & \cellcolor{cyan!20}{66.92$\pm$1.07} & {38.76$\pm$0.37} & {74.73$\pm$0.79} & {42.86$\pm$0.95} & \cellcolor{cyan!20}{81.00$\pm$0.70} & {34.00$\pm$0.30} & {69.39$\pm$0.30} & {36.74$\pm$0.30} \\
& KnowPO (w/o Aligned) & {54.45$\pm$0.51} & \cellcolor{cyan!20}{43.45$\pm$0.62} & {71.50$\pm$0.71} & \cellcolor{cyan!20}{43.00$\pm$1.07} & {48.29$\pm$0.93} & \cellcolor{cyan!20}{45.17$\pm$0.37} & {61.36$\pm$0.94} & \cellcolor{cyan!20}{50.30$\pm$0.94} & {70.00$\pm$0.93} & \cellcolor{cyan!20}{42.50$\pm$0.70} & {50.71$\pm$0.70} & \cellcolor{cyan!20}{46.27$\pm$0.48} \\
\hline
\end{tabular}}
\end{table*}

In this section, we conduct a series of experiments on two base models to answer the following research questions:
\begin{itemize}[leftmargin=*]
    \item \textbf{RQ1} (Section B): Does KnowPO outperform other approaches for resolving knowledge conflict across various base models and datasets?
    \item \textbf{RQ2} (Section C): What impact does each component has on the overall performance?
    \item \textbf{RQ3} (Section D): How does KnowPO alter the way LLMs utilize parametric knowledge?
    \item \textbf{RQ4} (Section E): How sensitive is KnowPO to hyper-parameters data ratio $\mathcal{R}_{error}$?
    \item \textbf{RQ5} (Section F): Does KnowPO training conducted in general domains remain effective in out-of-distribution (OOD) scenarios?
\end{itemize}

\subsection{A. Experimental Setup}
\label{Experimental Setup}
\paragraph{\textbf{Datasets }} We constructed the KnowPO training dataset based on SQuAD 2.0~\cite{rajpurkar2018knowdontknowunanswerable}. The test datasets comprise the following three types: (1) \textbf{SQuAD 2.0-Eval}, a validation set partitioned using the same construction method. (2) \textbf{Open-source counterfactual datasets:} RGB~\cite{chen2023benchmarkinglargelanguagemodels} and KNOT~\cite{liu2024untangleknotinterweavingconflicting} are two general-domain QA datasets containing counterfactual knowledge and contexts. We augmented these datasets with irrelevant contexts for testing purposes. Notably, RGB is a Chinese dataset. (3) \textbf{Domain-specific dataset:} CMB~\cite{cmb} is a multi-task QA dataset in the medical domain, encompassing 269,359 questions across four clinical medicine specialties of physicians, nurses, medical technicians, and pharmacists. Due to quantity constraints, we randomly sample 4,000 questions for testing. 

\paragraph{\textbf{Compared Methods }}
In order to explore the advantages of the KnowPO, we compare the KnowPO results against five other models: 
(1) \textbf{Base Model (Base)} answers user questions based on supplementary external knowledge, which can be considered as fundamental retrieve-and-read framework in RAG~\cite{lewis2021retrievalaugmented}. We selected Baichuan2-7B-chat~\cite{baichuan} and Llama2-13B-chat~\cite{touvron2023llama2openfoundation} as the base model and explored the gains brought by KnowPO: 
(2) \textbf{Naive Prompt-based Method (Prompt)} employs meticulously designed prompts to enhance the model's capability to adhere to external knowledge~\cite{zhou2023contextfaithfulpromptinglargelanguage}.
(3) \textbf{Advanced Prompt-based Method}: Chain of Thought (COT)~\cite{wei2023chainofthought} is a common method to enhance the performance of LLMs in downstream tasks. COT-VE~\cite{zhao2023verifyandeditknowledgeenhancedchainofthoughtframework} extends COT by guiding LLM to identify conflicting knowledge and modify its responses accordingly.
(4) \textbf{Finetuning}: KAFT~\cite{li2022largelanguagemodelscontrollable}  employs instruction fine-tuning to improve the LLM's adherence to contexts of varying relevance.
(5) \textbf{Decode-Based Method}: CAD~\cite{shi2023trustingevidencehallucinatecontextaware} uses a contrastive decoding-like method to adjust the probabilities of output tokens.
 Detailed settings of baselines can be reffered to in Appendix B.

\paragraph{\textbf{Metrics }}
We designed statistical metrics to evaluate the two capabilities of LLMs. For adherence capability, we utilized the conflicting contexts from the test set as supplementary knowledge, measuring the proportion $R_{Ad}$ of LLM responses that align with the conflicting knowledge within these contexts. For the RGB and KNOT datasets, the conflicting knowledge exclusively consists of counterfactual knowledge. For noise robustness, we employed the irrelevant contexts from the test set as supplementary knowledge, examining the proportion $R_{Ro}$ of LLM responses that correspond with the model's parameter knowledge.

\subsection{B. Performance Comparison(RQ 1)}
\label{Performance Comparison}
To answer RQ1, we conduct experiments and report results of the two metrics on Squad2.0-Eval, RGB and KNOT with two LLM turbos, as illustrated in Table \ref{tab:comparison}. From the reported results, we can find the following observations:

\textbf{Comparison of Baseline Methods and Base LLMs.} 
Through comparison, we observe that the KAFT method, fine-tuned with instructions, consistently outperforms across all experimental groups. This superior performance is primarily attributed to the use of contexts with varying degrees of relevance during fine-tuning, which significantly enhances the LLM's ability to focus on pertinent data while filtering out noise. In contrast, methods relying on the LLM's inherent capabilities for single or multiple interactions, such as Prompt or COT, tend to indiscriminately depend on external knowledge due to the LLM's limited noise recognition ability, leading to an increase in $R_{Ad}$, but a sharp decline in $R_{Ro}$. Particularly, COT-VE introduces additional noise by incorporating external knowledge for verification and editing , further complicating the model's ability to discern relevant information. As for CAD, as noted in related research, the contrastive decoding strategy compromises response coherence and utility, performing well on simple datasets like RGB but failing on more complex ones like Squad2.0-Eval and KNOT, thereby losing its practical value.

\textbf{Comparison of KnowPO and other methods.} Firstly, it is evident that our mothed, KnowPO, outperforms the baseline methods across all metrics.  For instance, the $R_{Ad}$ and $R_{Ro}$ scores see an improvement of approximately \textbf{37.07\%-125.06\%} and \textbf{80.91\%-416.93\%} for the Squad2.0-Eval  dataset with Baichuan2-7B-Chat. Moreover, compared to KAFT, best model in baselines, KnowPO uses more complicated contexts and comprehensive negative signals to enhance LLM's adherence capability and noise robustness.

\subsection{C. Ablation Study(RQ 2)}
To answer RQ2, we perform ablation studies to verify the effectiveness of KnowPO, as illustrated in Table \ref{tab:comparison}. Our observation can be summarized as follows:

\textbf{Effect of training phase.} Both the SFT and DPO phases positively contribute to enhancing the adherence capability and noise robustness of LLMs. Additionally, the preference learning method incorporating negative signals slightly outperforms SFT in improving the LLM's ability to utilize external knowledge, demonstrating the effectiveness of both training approaches.

\textbf{Effect of length imbalance.}
When data length are not aligned, we observe a significant impact of length bias, which slightly enhances $R_{Ro}$ but substantially reduces $R_{Ad}$. This is due to the model's inherent tendency to generate more verbose parametric answers, while the conflict answers derived through dataset construction are relatively short. Consequently, the model develops a preference for generating longer responses. Without length alignment between conflict and parametric answers, the model tends to consistently rely on parametric answers, thereby neglecting external knowledge and disrupting the balance between adherence capability and noise robustness.

\begin{figure}[htb]
  \centering
 \vspace{-0.25cm}
  \includegraphics[scale=0.3255]{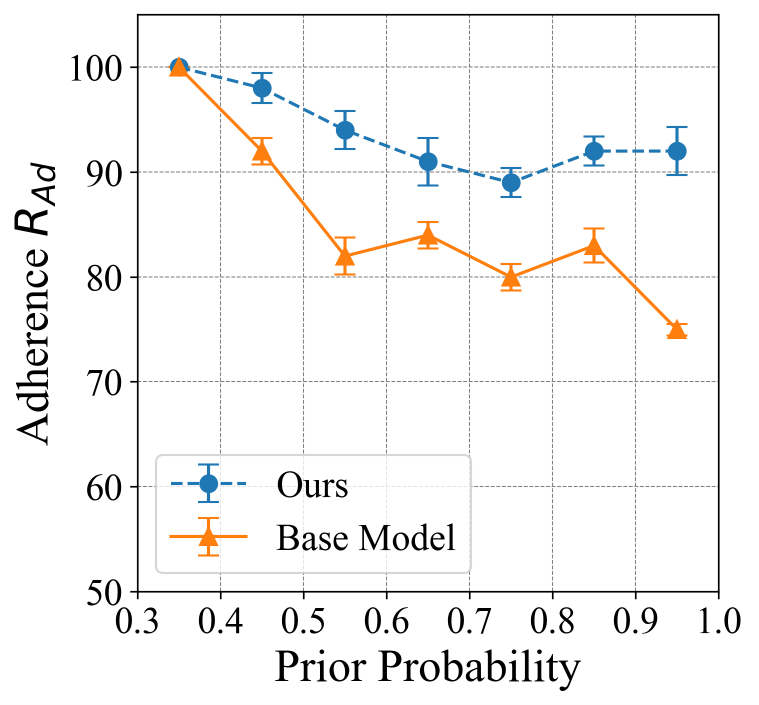}
  \includegraphics[scale=0.3255]{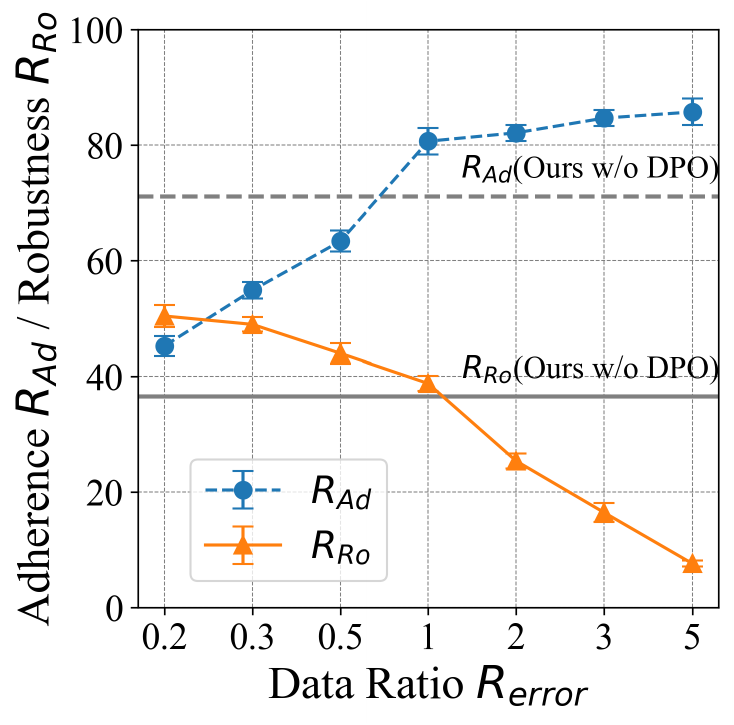}
  \vspace{-0.25cm}
  \caption{(\textbf{Left}.) The adherence ratio of LLM when encountering conflicting context with different prior probability on dataset RGB with Baichuan2-7B-Chat. (\textbf{Right}.) Hyper-parameter study with data ratio $\mathcal{R}_{error}$ on Squad2.0-Eval with Baichuan2-7B-Chat.
  }
   \label{fig:hyper_study}
   \vspace{-0.4cm}
\end{figure}

\subsection{D. Model Prior Analyse(RQ 3)}
The LLM's confidence in its responses is one of the factors influencing whether it prefers internal or external knowledge~\cite{wu2024clashevalquantifyingtugofwarllms}. We recorded the LLM's prior probability for parameter knowledge on the RGB dataset and measured the proportion of instances in each prior probability interval where the LLM followed external knowledge when encountering conflicting context. The model's prior response probability is computed from the average log probability of the response tokens without external knowledge. The results in Figure \ref{fig:hyper_study} show that, for base LLM, there is a general negative correlation between the prior probability of an answer and the proportion of following external knowledge; that is, the higher the prior probability, the less likely the answer is to be altered. However, after fine-tuning with KnowPO, although the overall trend remains negatively correlated, the trend is significantly mitigated, indicating that our method effectively enhances the LLM's adherence to external knowledge.

\subsection{E. Hyper-parameter Study(RQ 4)}
As analyzed in Section III.E, two types of preference pairs exhibit distinctly opposite behavioral tendencies: those simulating error contextual ignorance in conflicting contexts and those simulating error contextual overinclusion in irrelevant contexts. We conducted a series of analyses by adjusting the ratio $R_{error}$ between these two types of preference pairs from the list $[0.2, 0.3, 0.5, 1, 2, 3, 5]$. The results in Figure \ref{fig:hyper_study} indicate that as the proportion of the first type of preference pairs increases, the LLM becomes more inclined to utilize contextual knowledge, enhancing its adherence capability but also becoming more susceptible to noise, which in turn reduces its noise robustness.  Conversely, as the proportion of the second type increases, the LLM tends to disregard contextual information and respond directly, resulting in reduced $R_{Ad}$ but improved $R_{Ro}$. Notably, as the ratio $R_{error}$ increases from 1, the rate of improvement in adherence slows, while the decline in robustness becomes more pronounced. When the ratio $R_{error}$ decreases from 1, the curvature of $R_{Ad}$ and $R_{Ro}$ also shows the opposite trend. Based on these findings, we ultimately selected $R_{error} = 1$ as the optimal ratio, ensuring balanced improvements in both capabilities compared to SFT training.

\begin{table}[h]
\centering
\setlength{\tabcolsep}{2mm}  
\resizebox{\linewidth}{!}{
\begin{tabular}{c|cc|cc}
\hline
\textbf{} & \multicolumn{2}{c|}{\textbf{Baichuan2-7B-Chat}} & \multicolumn{2}{c}{\textbf{Llama2-13B-Chat}} \\ \cline{2-5} 
\textbf{} & {$R_{Ad}$} & {$R_{Ro}$} & {$R_{Ad}$} & {$R_{Ro}$} \\ \hline
\textbf{Base} & {58.69$\pm$0.79} & {10.66$\pm$0.64} & {60.95$\pm$0.59} & {8.21$\pm$0.44} \\ \hline
\textbf{KnowPO(w/o DPO)} & {95.66$\pm$0.48} & {23.70$\pm$0.32} & {83.88$\pm$0.40} & {16.53$\pm$0.82} \\ \hline
\textbf{KnowPO} & \cellcolor{red!20}{96.23$\pm$0.63} & \cellcolor{red!20}{24.12$\pm$0.51} & \cellcolor{red!20}{87.46$\pm$0.33} & \cellcolor{red!20}{21.24$\pm$0.47} \\ \hline
\end{tabular}}
\vspace{-0.15cm}
\caption{Performance comparison (in percent) on CMB}
\label{tab:ood}
\end{table}

\subsection{F. Generalization Analysis(RQ 5)}
To demonstrate the robust generalization capability of our method beyond the general domain training set, we conducted experiments on the CMB medical test set. By using medical triplets and documents to construct supplementary contexts in the medical domain, we created a conflict dataset for 4,000 CMB questions. The experimental results, as shown in Table \ref{tab:ood}, confirm that the model trained with KnowPO in the general domain also effectively enhances the LLM's adherence capability and noise robustness when transferred to the domain-specific context. The higher scores on CMB compared to those in Table \ref{tab:comparison} can be attributed to the fact that the contexts we constructed were less challenging than the real-world RAG knowledge.

\begin{table}[h]
\centering
\begin{tabular}{c|cc}
\hline
\textbf{Model} & \textbf{KnowPO(w/o DPO)} & \textbf{KnowPO} \\
\hline
Baichuan2-7B-Chat          & 3.65\%               & 4.70\%         \\
Llama2-13B-Chat         & 3.51\%               & 4.31\%         \\
\hline
\end{tabular}
\vspace{-0.15cm}
\caption{The match rate between LLM's parameter answers and conflicting answers after training.}
\label{tab:inject}
\end{table}

A potential risk of incorporating QA pairs and conflicting knowledge contexts into the training data is the inadvertent introduction of harmful information to the model. To assess whether the KnowPO-trained model retained any conflicting knowledge from the training set, we utilized prompts designed to extract LLM's parameter knowledge without providing supplemental knowledge. The results, presented in Table \ref{tab:inject}, indicate that the model exhibited virtually no retention of conflicting knowledge after the SFT and DPO phases of KnowPO. This finding corroborates that our method enhances the LLM's ability to leverage external knowledge rather than injecting specific knowledge into the model.

\section{V. Conclusion}
\label{Conclusion}
In this paper, we propose KnowPO, a \textbf{\underline{Know}}ledge-aware \textbf{\underline{P}}reference \textbf{\underline{O}}ptimization strategy to enhance LLM's adherence capability and noise robustness to external knowledge. 
Specifically, we abstract and simulate two common error types in scenarios with varying contextual relevance: Contextual Ignorance and Contextual Overinclusion. Based on instruction-tuning, we utilize negative gradient terms in the DPO comparative objectives to reduce the likelihood of undesired responses.
Furthermore, by aligning data lengths and balancing data ratios, we effectively mitigate preference imbalances inherent in DPO. Experimental evaluations across diverse datasets and two base models substantiate the efficacy and generalization capability of KnowPO. 
In the future, we will explore how the composition and proportion of different types of contexts affect the ability of LLMs to utilize external knowledge.

\bibliography{aaai25}

\newpage
\appendix
\section{A. Prompts used in KnowPO}
In this section, we provide a detailed display of all the prompts used. 

\begin{tcolorbox}[colback=lightgray!20,colframe=darkgray!80,title=Prompt B.1: Extract Parameter Answer]
This is a question about \{\textit{Title}\}. Please answer the question \{\textit{Question $q$}\}. Please provide a direct answer without analysis. If you are unsure or do not know the answer, please respond with `I don't know'.
\end{tcolorbox}

\begin{tcolorbox}[colback=lightgray!20,colframe=darkgray!80,title=Prompt B.2: Generate Counterfactual Answer]
Please generate speciously plausible but incorrect answer to the question. Provide only the false answers; do not reiterate the queries.

Question: What is the capital of France? Answer: Paris. Fake answer: Lyon.

Question: What is the highest mountain in the world? Answer: Mount Everest. Fake answer: Lhotse.

\textit{\ldots 7 more examples \ldots}

Question: Who is the founder of Microsoft? Answer: Bill Gates. Fake answer: Steve Jobs.

Question: \{\textit{Question $q$}\} Answer: \{\textit{Realistic Answer $a_{real}$}\} Fake answer:
\end{tcolorbox}

\begin{tcolorbox}[colback=lightgray!20,colframe=darkgray!80,title=Prompt B.3: Generate Contextual Ignorance]
This is a question about \{\textit{Title}\}. Please answer the question \{\textit{Question $q$}\}. Please provide a direct answer without analysis. If you are unsure or do not know the answer, please respond with `I don't know'.
\end{tcolorbox}

\begin{tcolorbox}[colback=lightgray!20,colframe=darkgray!80,title=Prompt B.4: Generate Contextual Overinclusion]
Please select a word from the provided context as an alternative answer to this question.

Question: \{\textit{Question $q$}\}

Potential answer: \{\textit{Conflicting Answer $a_{cf}$}\}

Context: \{\textit{Context $\tau$}\}\\

Please follow these requirements:

1. The answer must not be the same as the potential answer.

2. The alternative answer does not need to be correct, but it must appear in the context.

3. The alternative answer must be in a form that can answer the question and should be as reasonable as possible.
\end{tcolorbox}

\begin{tcolorbox}[colback=lightgray!20,colframe=darkgray!80,title=Prompt B.5: Instruction Tuning]
\textbf{[Instruction]} \quad As a knowledge-based QA expert, you will provide professional responses based on user's question, utilizing any supplemental knowledge provided to enhance the quality of your response. If the supplemental information is irrelevant to the question, rely on your own expertise to formulate an answer. If you are unsure about the answer, please respond with `I don't know'.

\textbf{[Supplemental Knowledge]} \quad \{\textit{Context $\tau$}\}

\textbf{[User's Question]} \quad \{\textit{Question $q$}\}

\textbf{[Answer]}
\end{tcolorbox}

\section{B. Further Experiment Details}

\subsection{B.1 Datasets Detail}
We employ four benchmark datasets for evaluation after additionally processing to introduce noisy context.

(1) SQuAD2.0 ~\cite{rajpurkar2018knowdontknowunanswerable} is 
a reading comprehension dataset encompassing multiple general domains, consisting of questions posed by crowdworkers on a set of Wikipedia articles, where the answer to every question is a segment of text, or span, from the corresponding reading passage, or the question might be unanswerable. We constructed conflicting knowledge and context as illustrated in Section III, and formulated irrelavant context based on human notations.

(2) RGB~\cite{chen2023benchmarkinglargelanguagemodels} is a corpus for RAG evaluation in both English and Chinese. To examine LLM's ability on noise robustness and counterfactual robustness, RGB provides QA pairs with counterfactual knowledge and context as well as noise context. We formulate testing dataset with more complex contexts based on the original corpus.

(3) KNOT~\cite{liu2024untangleknotinterweavingconflicting} is a corpus specially designed for knowledge conflict resolution examination in the form of question answering. KNOT categorizes different conflicting knowledge into three levels of difficulty, and we added noise context to this basis for performance testing.

(4) CMB~\cite{cmb} is a medical open-source query dataset, which are designed for multi-task Q\&A, encompass single and multiple-choice questions in the medical field. The CMB dataset utilizes qualifying exams as a data source in the four clinical medicine specialties of physicians, nurses, medical technicians, and pharmacists, with a total of 269,359 questions. Given the extensive size of the CMB dataset, we randomly sample 4,000 questions for testing. We constructed conflicting knowledge and context with original questions and answers, as well as irrelevant contexts. We do not require the LLM to complete multiple-choice questions but to directly answer the questions.

Details statistics of datasets can be found in Table \ref{tb:dataset statistics}.

\begin{table}[t]
  \centering
  \small
   \setlength\tabcolsep{2pt}
   \setlength{\abovecaptionskip}{0.1cm}
    \begin{threeparttable}
\begin{tabular}{c|ccc}
    \hline 
    \textbf{Dataset}& $Num_{conflicting}$ &  $Num_{irrelevant}$ & $Num_{total}$ \\ 
\hline
     SQuAD2.0-Train  & 11,000 & 7,700 & 18,700  \\
     SQuAD2.0-Eval  & 2,737 & 1,908 & 4,645   \\
     RGB & 200 & 200 & 400   \\
     KNOT  & 3,500 & 2,000 & 5,500   \\
     CMB  & 2,500 & 1,500 & 4000   \\
\hline
\end{tabular}
    \end{threeparttable}
  \caption{The statistics of datasets.}
  \label{tb:dataset statistics}%
\end{table}

\subsection{B.2 Baselines Detail}
In this section, we present the details of baselines.
\begin{itemize}[leftmargin=*]
    \item \textbf{Baichuan2-7B-chat}~\cite{baichuan}: The baseline refers to directly using the context provided in the constructed dataset to prompt the original Baichuan2-7B-Chat model to generate answers. Baichuan-7B-Chat is a multi-turn dialogue model fine-tuned on private dialogue instruction data, with a parameter scale of 7B, and the training data consists of private multi-turn dialogue instructions. \\ 
\end{itemize}
\begin{itemize}[leftmargin=*]
    \item \textbf{Llama2-13B-chat}~\cite{touvron2023llama2openfoundation}: The baseline refers to directly using the context provided in the constructed dataset to prompt the original Llama2-13B-Chat model to generate answers. Llama-2-13b-chat is developed and open-sourced by Meta AI. It performs excellently in tasks involving encoding, reasoning, and knowledge application, with a parameter scale of 13B. \\ 
\end{itemize}
\begin{itemize}[leftmargin=*]
    \item \textbf{CFP}~\cite{zhou2023context}: CFP designed two key strategies: opinion-based prompts and counterfactual demonstrations. Opinion-based prompts transform questions into inquiries about the narrator's opinion, forcing the model to pay more attention to the context. Counterfactual demonstrations, on the other hand, provide examples with incorrect facts to help the model enhance its reliance on context in situations of knowledge conflict. Experimental results show that these methods significantly improve the model's faithfulness across various tasks, particularly excelling in scenarios involving knowledge conflicts and the need for prediction abstention. \\ 
\end{itemize}
\begin{itemize}[leftmargin=*]
    \item \textbf{CoT}~\cite{wei2022chain}: CoT significantly improves the ability of large language models to handle complex reasoning tasks by generating a series of intermediate reasoning steps. This method does not require fine-tuning the model but instead guides the model to perform multi-step reasoning by providing input-output examples with reasoning chains in the prompts. Experiments show that Chain-of-Thought Prompting outperforms traditional prompting methods in arithmetic, commonsense, and symbolic reasoning tasks, particularly achieving new performance improvements in arithmetic benchmarks. \\ 
\end{itemize}
\begin{itemize}[leftmargin=*]
    \item \textbf{CoT-VE}~\cite{zhao2023verify}: CoT-VE aims to enhance the factual accuracy of predictions by post-editing reasoning chains. Specifically, this method generates verifying questions for uncertain reasoning instances and retrieves external knowledge to edit and correct erroneous information in the reasoning process. The approach has demonstrated significant performance improvements across multiple open-domain question-answering tasks, particularly excelling in tasks that involve complex reasoning. \\ 
\end{itemize}
\begin{itemize}[leftmargin=*]
    \item \textbf{CAD}~\cite{shi2023trusting}: CAD amplifies the difference in output probabilities with and without context, effectively reducing the influence of the model's prior knowledge on the generated results. This method requires no additional training and can significantly improve the factuality of the generated text, particularly excelling in tasks where the context conflicts with the model's prior knowledge. Experimental results show that CAD markedly enhances both the quality and factuality of generated outputs in summarization and knowledge conflict tasks across various language models. \\ 
\end{itemize}
\begin{itemize}[leftmargin=*]
    \item \textbf{KAFT}~\cite{li2022largelanguagemodelscontrollable}: KAFT constructs specific instruction-tuning datasets to enhance models' ability to use external knowledge by creating challenging counterfactual knowledge from training datasets and incorporating irrelevant knowledge to boost noise resistance. \\ 
\end{itemize}

\begin{table}[h]
\centering
\begin{tabular}{l l}
\hline
\textbf{Baseline} & \textbf{Code Repo URL} \\
\hline
CFP & \url{https://github.com/wzhouad/context-faithful-llm} \\
CoT-VE & \url{https://github.com/RuochenZhao/Verifyand-Edit} \\
CAD & \url{https://github.com/hongshi97/CAD} \\
\hline
\end{tabular}
\caption{Baseline Code URLs of Github Repository}
\label{tab:repo}
\end{table}

\subsection{B.3 Implementation Detail}
Implementations are done using the PyTorch 2.3.0 framework~\cite{paszke2019pytorch} in Python 3.10.14, on an Ubuntu server equipped with 8 NVIDIA V100 GPU and an Intel(R) Xeon(R) CPU.

For CFP, we tested the performance of the method under the knowledge conflict setting and adopted the strategy of opinion-based prompts, utilizing a zero-shot prompting paradigm in context learning.For CoT, we used a few-shot prompting paradigm to teach the LLM to generate reasoning chains. Since the context constructed in our dataset closely simulates the context retrieved in real RAG scenarios, for CoT-VE, after generating verification questions for uncertain reasoning instances, we directly used the constructed context as the retrieval result to prompt the LLM to generate answers to the verification questions, thereby editing the current reasoning instance. For KAFT, we constructed a conflicting dataset on Squad2.0 based on the descriptions in its paper, and conducted instruction fine-tuning using the templates employed in the original method with LoRA tuning. We set the learning rate to 1e-5 and training epoch to 1. For CAD, we strictly followed the original method's settings.
For KnowPO, we use lora in DPO stage and set the learning rate to 5e-6 and training epoch to 1. Besides, we replicated CFP, CoT-VE and CAD by referencing the officially released code in Table \ref{tab:repo}.

\end{document}